\documentclass{article}
\usepackage{spconf,amsmath,graphicx}

\usepackage{booktabs}
\usepackage{romannum}
\usepackage{enumitem}
\usepackage{amssymb}
\usepackage{multirow}
\usepackage{array}
\newcolumntype{P}[1]{>{\centering\arraybackslash}p{#1}}
\usepackage{hyperref}
\usepackage{xcolor}

\title{CMTM: Cross-Modal Token Modulation\\for Unsupervised Video Object Segmentation}

\name{Inseok Jeon\quad Suhwan Cho\quad Minhyeok Lee\quad Seunghoon Lee\quad Minseok Kang \vspace{-5mm}}

\address{\textit{Jungho Lee\quad Chaewon Park\quad Donghyeong Kim\quad Sangyoun Lee}\\ \\
Yonsei University, Republic of Korea}

\begin{document}

\maketitle

\begin{abstract}
Recent advances in unsupervised video object segmentation have highlighted the potential of two-stream architectures that integrate appearance and motion cues. However, fully leveraging these complementary sources of information requires effectively modeling their interdependencies. In this paper, we introduce cross-modality token modulation, a novel approach designed to strengthen the interaction between appearance and motion cues. Our method establishes dense connections between tokens from each modality, enabling efficient intra-modal and inter-modal information propagation through relation transformer blocks. To improve learning efficiency, we incorporate a token masking strategy that addresses the limitations of relying solely on increased model complexity. Our approach achieves state-of-the-art performance across all public benchmarks, outperforming existing methods. The code is released on \textit{\textcolor{blue}{\url{https://github.com/InSeokJeon/CMTM}}}
\end{abstract}


\vspace{-3mm}
\section{Introduction}
\vspace{-3mm}
Video object segmentation is a critical task in computer vision that aims to accurately segment objects at the pixel level in video sequences. Methods for video object segmentation can generally be classified based on the availability of guidance for target identification. In semi-supervised video object segmentation, a segmentation mask of the target object is provided in the first frame, which serves as a reference for tracking and segmenting the object throughout the video. In contrast, unsupervised video object segmentation (UVOS) requires models to automatically detect and segment salient objects across the video without any external guidance.

Recent progress in UVOS has demonstrated the potential of two-stream architectures that combine appearance and motion cues. Appearance cues provide valuable visual information, such as color and texture, while motion cues capture object movement across frames. These complementary sources of information, when effectively integrated, can significantly enhance segmentation performance. However, fully exploiting the potential of both appearance and motion cues requires effective modeling of their interdependencies.

Existing UVOS methods, while advancing the field, often face challenges in fully leveraging these cues. One limitation is that many methods rely on complex encoder architectures but fail to explicitly guide the model in learning meaningful intra-modal representations within each modality. This leads to noisy or incomplete features, limiting the integration of appearance and motion information and ultimately constraining the model's ability to perform effectively. Another key limitation is the lack of robust mechanisms for inter-modal relation reasoning, which prevents the model from understanding how the appearance and motion modalities complement each other. Directly combining features from both modalities without properly modeling their relationships can result in an imbalanced integration, where irrelevant or redundant information from one modality may obscure important contributions from the other.

To address these limitations, we argue that effective UVOS requires two key components: 1) enhanced intra-modal representations and 2) robust inter-modal relation reasoning. To achieve this, we introduce cross-modality token modulation (CMTM), a novel framework designed to improve the interaction between appearance and motion cues. CMTM utilizes dense transformer blocks to enhance intra-modal representations and capture meaningful inter-modal interactions. Furthermore, we introduce a token masking strategy to facilitate the effective learning of these dense transformer blocks. By employing this masking strategy during training, the model learns to optimize both spatial and inter-modal interaction modeling, ensuring the effective integration of appearance and motion features. We evaluate our approach on public benchmark datasets, outperforming existing approaches by a significant margin.


\noindent Our main contributions can be summarized as follows:

\begin{itemize}[itemsep=0.4em, parsep=-2pt, topsep=-2pt] 
\item We introduce cross-modality token modulation, a novel framework that enhances both intra- and inter-modal representations through dense transformer blocks. 
\item We propose a token masking that promotes the efficient learning of dense transformer blocks. 
\item Our method achieves state-of-the-art performance on standard UVOS benchmarks, showcasing its effectiveness across diverse scenarios. 
\end{itemize}

\vspace{-2mm}
\section{Related Work}
\vspace{-2mm}
\textbf{Unsupervised video object segmentation.} 
A central approach in UVOS is the integration of appearance and motion cues to accurately generate segmentation masks. Two-stream architectures that combine these cues are widely explored. MATNet~\cite{MATNET} introduces a two-stream encoder that merges RGB images with optical flow maps to enhance spatio-temporal representations. FSNet~\cite{FSNET} proposes a full-duplex strategy with a bi-directional interaction module to ensure mutual refinement between appearance and motion cues. Similarly, AMCNet~\cite{AMCNET} employs a co-attention gating mechanism for effective fusion of appearance and motion information. TransportNet~\cite{TRANSPORTNET} leverages optimal structural matching using a Sinkhorn layer, while RTNet~\cite{RTNET} introduces a reciprocal transformation network. HFAN~\cite{HFAN} presents a hierarchical feature alignment network that aligns features at multiple scales. GSANet~\cite{GSANET} employs a guided slot attention mechanism to reinforce spatial structural information.

\begin{figure}[t!]
\centering
\includegraphics[width=1\linewidth]{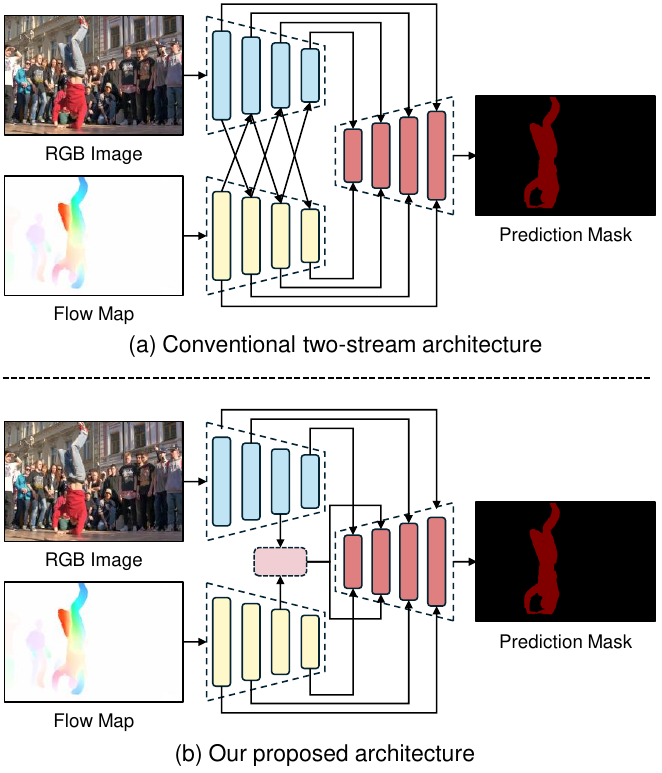}
\vspace{-8mm}
\caption{A visual comparison between the conventional two-stream architecture and our proposed architecture.}
\label{figure1}
\end{figure}

Despite these advancements, many existing methods rely on basic fusion techniques and fail to fully leverage intra- and inter-modal dependencies. In contrast, our framework emphasizes a deeper understanding of both intra- and inter-modal interactions through dense transformer blocks and a masked learning protocol. As shown in Fig.~\ref{figure1}, we provide a visual comparison between the conventional two-stream architecture and our proposed architecture.

\vspace{-2mm}
\section{Approach}
\vspace{-2mm}

\subsection{Task Formulation}
\vspace{-2mm}
In UVOS, the objective is to generate binary segmentation masks $M$ from each input video sequence. To this end, optical flow maps $F$ are first extracted from RGB images $I$, where 2-channel motion vectors are converted to 3-channel RGB values. Our method processes each frame independently, leveraging the corresponding image $I_i$ and flow map $F_i$ to predict the mask $M_i$, where $i$ indicates each video frame.

\vspace{-2mm}
\subsection{Overall Architecture}
\vspace{-2mm}
Our framework consists of three main components: two-stream encoders, a CMTM module, and a decoder. The two-stream encoders independently process the RGB image $I$ and the flow map $F$, allowing each encoder to extract modality-specific features, denoted as $\mathbf{F}_{app}$ and $\mathbf{F}_{mo}$, respectively. Positioned between the encoders and the decoder, the CMTM module employs dense transformer blocks to enhance intra-modal representations and enable robust inter-modal relation reasoning. Finally, the decoder refines these enhanced features to generate the binary segmentation masks.

\vspace{-2mm}
\subsection{Cross-Modality Token Modulation}
\begin{figure*}[t!]
    \centerline{\includegraphics[width=\textwidth]{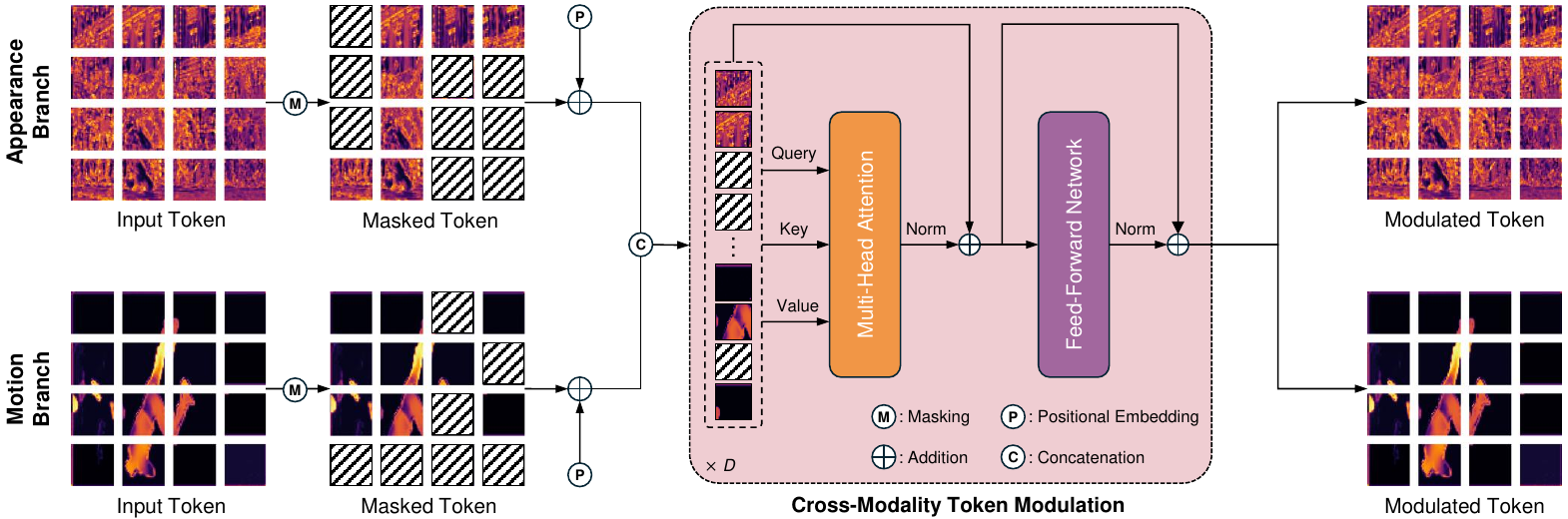}}
    \vspace{-2mm}
    \caption{Illustration of the CMTM module, where $D$ representing the number of dense transformer blocks involved in the process.}
    \vspace{-2mm}
    \label{figure2}
\end{figure*}

\vspace{-2mm}
Existing two-stream VOS methods often struggle with suboptimal intra-modal representations and inadequate inter-modal relation reasoning, which ultimately undermines the quality and reliability of predictions. To address these challenges, we propose CMTM, a method that enhances intra-modal embeddings and facilitates robust inter-modal interactions through dense transformer blocks and a token masking strategy. A visual illustration of the CMTM is provided in Fig.~\ref{figure2}.

\vspace{0.1cm}
\noindent \textbf{Dense transformer block.}
To effectively combine complementary information from appearance and motion cues, we utilize dense relation modeling through a self-attention mechanism. This approach simultaneously refines intra-modal representations and captures inter-modal interactions, ensuring robust feature integration.

The module takes feature maps $\mathbf{F}_{app} \in \mathbb{R}^{H \times W \times C}$ from the appearance encoder and $\mathbf{F}_{mo} \in \mathbb{R}^{H \times W \times C}$ from the motion encoder as input. These feature maps are tokenized into $\mathbf{T}_{app} \in \mathbb{R}^{N \times C}$ and $\mathbf{T}_{mo} \in \mathbb{R}^{N \times C}$, where $N$ denotes the total number of tokens obtained by flattening the spatial dimensions of the feature maps. The appearance and motion tokens are then concatenated along the token dimension as $[\mathbf{T}_{app}^{m}; \mathbf{T}_{mo}^{m}]$, where $\mathbf{T}_{app}^{m}$ and $\mathbf{T}_{mo}^{m}$ represent the masked tokens. This concatenation allows the model to jointly process appearance and motion modalities, exploiting their complementary characteristics for robust feature integration. The self-attention mechanism then processes these concatenated tokens to capture both intra-modal refinements and inter-modal interactions. The self-attention operation in our approach is formulated as:

\begin{align}
\mathbf{A} &= \text{Softmax}\left(\frac{\mathbf{Q}\mathbf{K}^\top}{\sqrt{C}}\right)\mathbf{V}\\
           &= \text{Softmax}\left(\frac{[\mathbf{Q}_{\text{app}}^{m}; \mathbf{Q}_{\text{mo}}^{m}]
                [\mathbf{K}_{\text{app}}^{m}; \mathbf{K}_{\text{mo}}^{m}]^\top}{\sqrt{C}}\right)
                [\mathbf{V}_{\text{app}}^{m}; \mathbf{V}_{\text{mo}}^{m}] \nonumber
~,
\end{align}
where $\mathbf{Q}$, $\mathbf{K}$, and $\mathbf{V}$ are the query, key, and value matrices derived from the concatenated tokens. This process can also be written as:
\begin{align}
\mathbf{A} =
\begin{bmatrix}
\mathbf{W}_{\text{app,app}} \mathbf{V}_{\text{app}} + \mathbf{W}_{\text{app,mo}} \mathbf{V}_{\text{mo}} \\
\mathbf{W}_{\text{mo,app}} \mathbf{V}_{\text{app}} + \mathbf{W}_{\text{mo,mo}} \mathbf{V}_{\text{mo}}
\end{bmatrix}~.
\end{align}
This formulation enables dense relation modeling by leveraging pixel-level tokens to integrate both spatial and semantic information from appearance and motion features. Specifically, $\mathbf{W}_{\text{app,app}} \mathbf{V}_{\text{app}}$ facilitates intra-modal feature extraction within the appearance domain, while $\mathbf{W}_{\text{app,mo}} \mathbf{V}_{\text{mo}}$ enriches the appearance features with motion information. Similarly, motion tokens undergo the same token modulation process. Through the use of dense transformer blocks, we achieve both intra-modal representation enhancement and inter-modal representation modulation.

\vspace{0.1cm}
\noindent\textbf{Token masking.}
Simply increasing model complexity with dense transformer blocks does not inherently guarantee improved performance. Instead, it often results in models prone to overfitting or unable to generalize effectively across diverse scenarios. Without explicit mechanisms to guide the learning process, increased complexity may amplify irrelevant or noisy features, hindering the extraction of meaningful semantic patterns. To address these challenges, our framework employs a masking strategy that enhances the model's focus on critical semantic cues, fostering a more comprehensive understanding of the input.

In this approach, random masking is applied to the input tokens of the CMTM module. A binary mask $\mathbf{M} \in \{0, 1\}^N$ is used for token filtering, selecting a pre-defined number of tokens from $\mathbf{T}$ based on the masking ratio $\Psi$. Elements with a value of 1 in $\mathbf{M}$ denote the retained tokens. The masking process is formally expressed as:
\begin{align} 
\mathbf{T}^{m} &= \mathbf{T} \odot \mathbf{M}~, 
\end{align}
where $\odot$ represents the Hadamard product. This process is independently applied to both appearance and motion tokens. The binary mask $\mathbf{M}$ is randomly generated at each training iteration, ensuring diverse masking patterns and preventing overfitting to specific token positions. Masked tokens are replaced with learnable mask tokens initialized as trainable parameters, encouraging the model to infer missing information by leveraging contextual relationships from unmasked tokens, thus promoting robust feature learning.

By integrating this masking strategy, the model avoids over-reliance on specific features and develops a deeper understanding of semantic dependencies. This facilitates effective training of the dense transformer blocks, enabling them to capture complex patterns and significantly enhance the framework's overall performance. Note that the masking process is exclusively applied during the training stage of the network.

\vspace{-2mm}
\subsection{Decoding}
\vspace{-2mm}
The decoding process is based on the structure of our baseline model, FakeFlow~\cite{FAKEFLOW}. However, unlike FakeFlow, the third encoder's features are modulated using the CMTM module. Consequently, the decoder utilizes the modulated tokens from the CMTM module instead of directly using the features from the encoders. Apart from this modification, all other settings remain identical. The decoder takes multi-resolution features as input and progressively decodes them to produce the final binary segmentation mask.

\begin{table*}[t!]
\centering 
\small
\caption{Quantitative evaluation on the DAVIS 2016 validation set, FBMS test set and YouTube-Objects dataset. OF and PP indicate the use of optical flows and post-processing techniques, respectively. $\ast$ denotes speed calculated on our hardware.}
\vspace{1mm}
\begin{tabular}{p{2cm}P{2cm}P{2.3cm}P{4mm}P{4mm}P{6mm}P{8mm}P{8mm}P{8mm}P{8mm}P{8mm}P{8mm}}
\toprule
\multicolumn{6}{c}{} &\multicolumn{3}{c}{DAVIS 2016} &\multicolumn{1}{c}{FBMS} &\multicolumn{1}{c}{YTO} &\multicolumn{1}{c}{LVD}\\
\cmidrule(lr){7-12}
Method &Publication &Backbone &OF &PP &fps &$\mathcal{G}_\mathcal{M}$ &$\mathcal{J}_\mathcal{M}$ &$\mathcal{F}_\mathcal{M}$ &$\mathcal{J}_\mathcal{M}$ &$\mathcal{J}_\mathcal{M}$ &$\mathcal{J}_\mathcal{M}$\\
\midrule
RTNet~\cite{RTNET}                &CVPR'21        &ResNet-101     &\checkmark    &\checkmark &-           &85.2 &85.6 &84.7 &-    &71.0 \\
FSNet~\cite{FSNET}                &ICCV'21        &ResNet-50~\cite{RESNET}      &\checkmark    &\checkmark &12.5        &83.3 &83.4 &83.1 &-    &-  &- \\
TransportNet~\cite{TRANSPORTNET}  &ICCV'21        &ResNet-101~\cite{RESNET}     &\checkmark    &           &12.5        &84.8 &84.5 &85.0 &78.7 &-  &- \\
AMC-Net~\cite{AMCNET}             &ICCV'21        &ResNet-101~\cite{RESNET}     &\checkmark    &\checkmark &17.5        &84.6 &84.5 &84.6 &76.5 &71.1 &- \\
D$^2$Conv3D~\cite{D2CONV3D}       &WACV'22        &ir-CSN-152~\cite{CSN}     &              &           &-           &86.0 &85.5 &86.5 &-    &- &- \\
IMP~\cite{IMP}                    &AAAI'22        &ResNet-50~\cite{RESNET}      &              &           &1.79        &85.6 &84.5 &86.7 &77.5 &- &- \\
HFAN~\cite{HFAN}                  &ECCV'22        &MiT-b2~\cite{SEGFORMER}         &\checkmark    &           &12.8$^\ast$ &87.5 &86.8 &88.2 &-    &73.4  &80.2\\
OAST~\cite{OAST}                  &ICCV'23        &MobileViT3D~\cite{MOBILEVIT}    &\checkmark    &           &-           &87.0 &86.6 &87.4 &83.0 &- &- \\
SimulFlow~\cite{SIMULFLOW}        &ACMMM'23       &MiT-b2~\cite{SEGFORMER}         &\checkmark    &           &25.2$^\ast$ &88.3 &87.1 &89.5 &84.1 &- &- \\
GFA~\cite{GFA}                    &AAAI'24        &-              &\checkmark    &           &-           &88.2 &87.4 &88.9 &82.4 &74.7 &- \\
GSA-Net~\cite{GSANET}             &CVPR'24        &MiT-b2~\cite{SEGFORMER}         &\checkmark    &           &38.2        &88.2 &87.4 &89.0 &82.3 &- &- \\
FakeFlow~\cite{FAKEFLOW}          &arXiv'24       &MiT-b2~\cite{SEGFORMER}         &\checkmark    &           &29.5$^\ast$ &88.5 &88.0 &89.0 &84.6 &\textbf{75.0} &80.6 \\
\midrule
\textbf{CMTM}                     &               &MiT-b2~\cite{SEGFORMER}         &\checkmark    &           &19.8$^\ast$ &\textbf{89.2} &\textbf{88.5} &\textbf{89.8} &\textbf{84.7} &74.7 &\textbf{80.8}\\
\bottomrule
\end{tabular}
\label{table1}
\vspace{-0.3cm}
\end{table*}

\vspace{-2mm}
\subsection{Implementation Details}
\vspace{-2mm}

\noindent\textbf{CMTM architecture.}
The proposed CMTM module is integrated into the third encoding layer of the appearance and motion streams. This design choice strikes a balance between spatial granularity and semantic richness, making it suitable for precise object delineation while maintaining computational efficiency. We utilized a fixed positional embedding and additionally introduced a Modality Embedding to differentiate appearance and motion tokens.

\vspace{0.1cm}
\noindent\textbf{Training strategy.}
To ensure a fair comparison with the baseline FakeFlow, we adhere to the same two-stage training protocol. In the first stage, the model is pre-trained on the YouTube-VOS 2018~\cite{YTVOS} training set, where all objects in each video sequence are merged into a single salient object. In the second stage, the network is fine-tuned using a combination of the DAVIS 2016~\cite{DAVIS} training set and the DUTSv2~\cite{DUTS, FAKEFLOW} dataset, with a mixing ratio of 1:3.

\vspace{0.1cm}
\noindent\textbf{Training details.}
For implementation, we adopt the MiT-b2 backbone~\cite{SEGFORMER} with an input resolution of $512 \times 512$ as the default configuration. Network optimization is performed using the cross-entropy loss function and the Adam optimizer~\cite{adam}, with a learning rate of $1e^{-5}$. All experiments are conducted using two GeForce RTX TITAN GPUs.

\vspace{-2mm}
\section{Experiment}
\vspace{-2mm}
We conduct extensive experiments to validate the effectiveness of our method. The evaluation datasets include the DAVIS 2016~\cite{DAVIS} validation set (D), the FBMS~\cite{FBMS} test set (F), the YouTube-Objects~\cite{YTOBJ} (Y), and Long-Videos~\cite{LVID} dataset (L). Speed evaluations are performed using a single GeForce RTX 2080 Ti GPU.

\vspace{-2mm}
\subsection{Evaluation Metrics}
\vspace{-2mm}
To evaluate the performance of our method, we use three metrics: region similarity $\mathcal{J}$, boundary accuracy $\mathcal{F}$, and their average $\mathcal{G}$. These metrics offer a comprehensive assessment by considering both the overlap and boundary alignment between the predicted and ground truth masks.

\vspace{-2mm}
\subsection{State-of-the-Art Comparison}
\vspace{-2mm}

\begin{table}[t!]
\centering 
\small
\vspace{-3mm}
\caption{Direct comparison with FakeFlow.}
\vspace{1mm}
\begin{tabular}{p{1.7cm}|P{1.3cm}|P{6mm}P{6mm}P{6mm}P{6mm}}
\toprule
Method &Backbone &D &F &Y &L\\
\midrule
\multirow{3}*{FakeFlow~\cite{FAKEFLOW}} &MiT-b0 &86.7 &81.2 &70.4 &75.7\\
&MiT-b1 &87.3 &81.8 &73.4 &77.4\\
&MiT-b2 &88.5 &84.6 &75.0 &80.6\\
\midrule
\multirow{3}*{CMTM} &MiT-b0 &87.8 &83.4 &70.8 &76.8\\
&MiT-b1 &88.7 &83.4 &73.2 &77.1\\
&MiT-b2 &89.2 &84.7 &74.7 &80.8\\
\bottomrule
\end{tabular}
\label{table2}
\vspace{-3mm}
\end{table}

\noindent\textbf{Quantitative results.}
Table~\ref{table1} provides a comparative analysis of our method against existing approaches across four benchmark datasets. Our method consistently achieves state-of-the-art performance, demonstrating its robustness and adaptability to diverse segmentation scenarios while maintaining an optimal balance between accuracy and inference speed. Table~\ref{table2} presents a direct comparison between our method and FakeFlow~\cite{FAKEFLOW} across different backbone versions, further highlighting the superiority of our approach.

\begin{figure*}[t!]
    \centerline{\includegraphics[width=\textwidth]{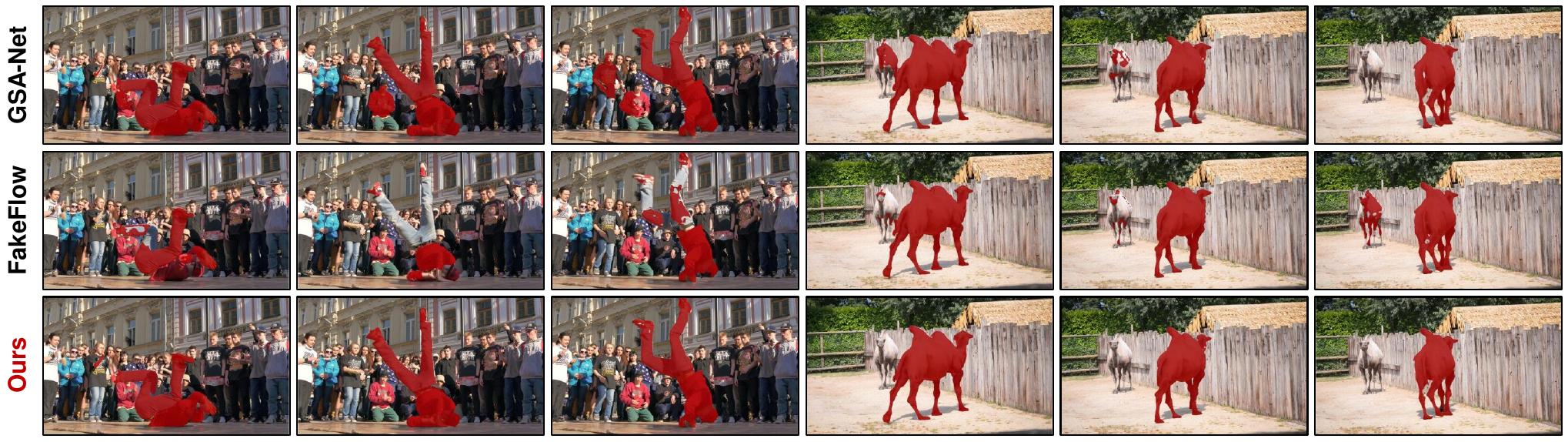}}
    \vspace{-3mm}
    \caption{Qualitative comparison of our method with state-of-the-art approaches.}
    \vspace{-4mm}
    \label{figure3}
\end{figure*}

\vspace{0.1cm} 
\noindent\textbf{Qualitative results.}
Fig.~\ref{figure3} presents qualitative comparisons between our method and current approaches. The visualizations underscore the superiority of our method in accurately segmenting object boundaries and preserving fine-grained details, even in challenging scenarios. 


\vspace{-2mm}
\subsection{Analysis}
\vspace{-2mm}
To validate the effectiveness and efficiency of the proposed CMTM, we conduct a thorough analysis. All ablation studies are performed using the MiT-b0 backbone and are restricted to stage 2 training.

\vspace{0.1cm}
\noindent\textbf{Effectiveness of CMTM.} 
As shown in Table~\ref{table3}, a comprehensive evaluation of the dense transformer block and token masking is presented. The results indicate that applying dense transformer blocks alone does not yield significant performance improvements, as increasing model complexity without an explicit learning protocol is ineffective. The dense transformer blocks show their efficacy when combined with token masking. Additionally, the two-stream application of CMTM outperforms the single-stream application, highlighting the effectiveness of intra-modal information propagation.

\vspace{0.1cm}
\noindent\textbf{Masking ratio.} 
Table~\ref{table4} presents the performance across different masking ratios. A ratio value of 0.0 represents the application of CMTM without token masking (i.e., using only dense transformer blocks). The highest performance is achieved with a masking ratio of 0.4.

\begin{table}[t!]
\small
\centering
\caption{Ablation study on the CMTM.}
\vspace{1mm}
\begin{tabular}{c|P{0.8cm}|P{0.8cm}|P{0.8cm}|P{0.6cm}P{0.6cm}P{0.6cm}}
\toprule
Version &App. &Mo. &Mask &D &F &Y\\ 
\midrule
\Romannum{1} &\checkmark & &           &85.9 &81.4 &69.4\\ 
\Romannum{2} &\checkmark & &\checkmark &86.2 &80.7 &69.7\\ 
\midrule
\Romannum{3} & &\checkmark &           &85.7 &81.2 &70.6\\ 
\Romannum{4} & &\checkmark &\checkmark &85.7 &77.9 &69.6\\ 
\midrule
\Romannum{5} &\checkmark &\checkmark & &86.9 &79.3 &68.2\\
\Romannum{6} &\checkmark &\checkmark &\checkmark &87.5 &79.9 &69.1 \\
\bottomrule
\end{tabular}
\vspace{-5mm}
\label{table3}
\end{table}

\begin{table}[t!]
\small
\centering
\caption{Ablation study on the Masking Ratio.}
\vspace{1mm}
\begin{tabular}{c|P{0.8cm}|P{0.6cm}P{0.6cm}P{0.6cm}}
\toprule
Version &$\Psi$ &D &F &Y\\ 
\midrule
\Romannum{1} &0.0 &86.9 &79.3 &68.2\\ 
\Romannum{2} &0.2 &87.0   &80.6 &69.2\\ 
\Romannum{3} &0.4 &87.5 &79.9 &69.1\\ 
\Romannum{4} &0.6 &86.2 &80.9 &68.9\\ 
\Romannum{5} &0.8 &85.6 &80.1 &68.7\\ 
\bottomrule
\end{tabular}
\label{table4}
\vspace{-0.5cm}
\end{table}

\begin{figure}[t!]
\centering
\vspace{2mm}
\includegraphics[width=1\linewidth]{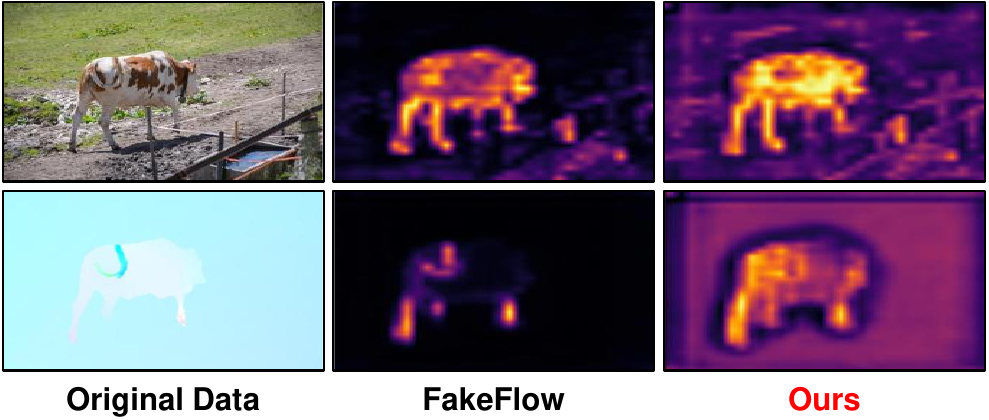}
\vspace{-7mm}
\caption{Comparison of the feature maps generated by FakeFlow and our proposed method CMTM.}
\vspace{-3mm}
\label{figure4}
\end{figure}

\begin{figure}[t!]
\centering
\vspace{2mm}
\includegraphics[width=1\linewidth]{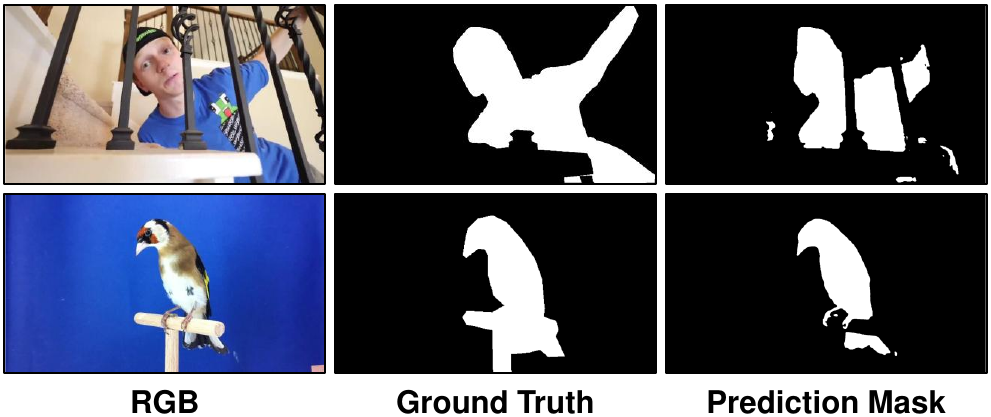}
\vspace{-7mm}
\caption{Qualitative evaluation of CMTM.}
\vspace{-3mm}
\label{figure5}
\end{figure}

\vspace{0.1cm}
\noindent\textbf{Learned feature visualization.} 
CMTM is designed to enhance the extraction of meaningful feature representations for primary object detection. To verify its effectiveness, we compare the visualized feature maps of our baseline, FakeFlow, and the proposed CMTM in Fig.~\ref{figure4}. By capturing both intra- and inter-modal relationships, CMTM effectively modulates encoder features, enriching feature representations. Its fusion of appearance and motion cues enables a clearer distinction of the primary object.

\vspace{0.1cm}
\noindent\textbf{Qualitative analysis.} 
Due to the annotation burden, some ground truth masks in UVOS datasets lack fine-grained details, as shown in Fig.~\ref{figure5}. Our proposed CMTM produces high-fidelity mask predictions for primary objects, even in severe occlusion scenarios, often surpassing the quality of the provided ground truth.

\vspace{-4mm}
\section{Conclusion}
\vspace{-2mm}
We introduce the cross-modality token modulation (CMTM) framework, which enhances unsupervised video object segmentation by integrating intra- and inter-modal relationships. CMTM outperforms state-of-the-art methods, demonstrating significant improvements in segmentation accuracy.

\vspace{2mm}
\fontsize{8pt}{8pt}\selectfont
\noindent\textbf{Acknowledgements.} 
This work was supported by the Korea Institute of Science and Technology (KIST) Institutional Program (Project No.2E33612-25-016), National Research Foundation of Korea (NRF) grant funded by the Korea government (MSIT)(No. RS-2024-00423362) and Yonsei Signature Research Cluster Program of 2025 (2025-22-0013).
\bibliographystyle{IEEEbib}

\fontsize{9.8pt}{9.8pt}\selectfont
\bibliography{strings,refs}

\end{document}